# MABL: Bi-Level Latent-Variable World Model for Sample-Efficient Multi-Agent Reinforcement Learning

Aravind Venugopal
Carnegie Mellon University
Pittsburgh, Pennsylvania, USA

Stephanie Milani
Carnegie Mellon University
Pittsburgh, Pennsylvania, USA

Fei Fang
Carnegie Mellon University
Pittsburgh, Pennsylvania, USA

Balaraman Ravindran
Indian Institute of Technology, Madras
Chennai, Tamil Nadu, India

## ABSTRACT

Multi-agent reinforcement learning (MARL) methods often suffer from high sample complexity, limiting their use in real-world problems where data is sparse or expensive to collect. Although latent-variable world models have been employed to address this issue by generating abundant synthetic data for MARL training, most of these models cannot encode vital global information available during training into their latent states, which hampers learning efficiency. The few exceptions that incorporate global information assume centralized execution of their learned policies, which is impractical in many applications with partial observability.

We propose a novel model-based MARL algorithm, MABL (Multi-Agent Bi-Level world model), that learns a bi-level latent-variable world model from high-dimensional inputs. Unlike existing models, MABL is capable of encoding essential global information into the latent states during training while guaranteeing the decentralized execution of learned policies. For each agent, MABL learns a global latent state at the upper level, which is used to inform the learning of an agent latent state at the lower level. During execution, agents exclusively use lower-level latent states and act independently. Crucially, MABL can be combined with any model-free MARL algorithm for policy learning. In our empirical evaluation with complex discrete and continuous multi-agent tasks including SMAC, Flatland, and MAMuJoCo, MABL surpasses SOTA multi-agent latent-variable world models in both sample efficiency and overall performance.

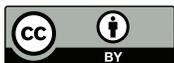

## 1 INTRODUCTION

Multi-agent reinforcement learning (MARL) offers a powerful, versatile approach for addressing a variety of real-world problems that require coordination among multiple agents, such as the control of robot swarms [1, 23], autonomous vehicles [6], and more [18, 44]. These scenarios offer myriad challenges: agents must often learn to behave from high-dimensional, partially observable inputs while grappling with the issue of non-stationarity induced by other agents simultaneously learning in the environment [22, 29]. The resulting complexity often translates to a tremendous amount of environment interactions for learning effective policies [9]. In practical scenarios, collecting such interaction data is resource-intensive and time-consuming [2], underscoring the importance of *sample efficiency*.

In the single-agent setting, model-based RL has shown promise in improving sample efficiency by enabling agents to utilize predictive models of environment dynamics [4, 35, 40]. These models, commonly referred to as *world models*, are often used to generate synthetic data from which agents can learn how to act. These approaches have been shown to improve sample efficiency by reducing the number of environment interactions needed to learn good behavior [14, 20, 25, 37]. However, they still require learning in a high-dimensional space. In recent years, however, model-based RL algorithms have employed latent-variable world models [12, 13, 20] to learn low-dimensional latent states from high-dimensional inputs. Trajectories of latent states generated by the model are then used for policy learning. Although this family of methods represents the state-of-the-art in the single-agent setting, they have only recently been brought to bear in MARL [41]. Yet, current approaches [7, 19] suffer from key limitations.

In MARL, the established paradigm of centralized training with decentralized execution (CTDE) [22, 32] offers a pragmatic balance that enables centralized learning by allowing agents access to additional information during training, as long as each agent only accesses its private observation during policy execution. This paradigm ensures scalability and practicality in scenarios where agents have to act in a decentralized manner. We refer to the totality of information available to an agent during training, including its private observation, as "global information". By following the CTDE paradigm, model-based MARL agents can utilize global information in their models during training to enhance latent representation learning, potentially leading to more sample-efficient learning.

However, existing multi-agent latent-variable world models are either incapable of incorporating global information [19] or do so, but fail to ensure that agents only use their own observations during execution [7, 42]. This is because they compute latent states by accessing the observations of all agents. Although the latter class of approaches yields state-of-the-art sample efficiency, they perform centralized training with centralized execution (CTCE). This paradigm requires a continuous transfer of inputs between agents and a centralized controller, a transformer block [7, 42], during execution. This, in turn, leads to increased latency, bottlenecks, and high susceptibility to communication failures. It also makes these approaches inapplicable to settings with partial observability and without communication channels between agents. Moreover, these approaches are not designed to incorporate any global information that is available in addition to agents' observations, and omitting such crucial information during training can be detrimental to learning.

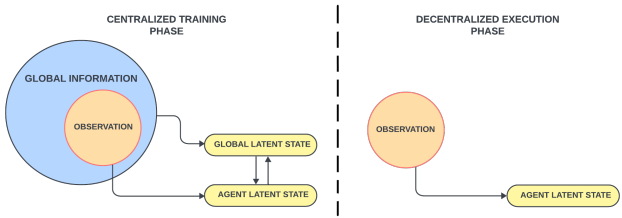

Figure 1: Overview of how the MABL bi-level model encodes global information into the global latent state while training while informing the agent latent state (left), which is computed from the agent's observation during execution (right).

To address the problem of incorporating global information while maintaining CTDE, we introduce *Multi-Agent Bi-Level World Model (MABL)*, the first CTDE multi-agent latent-variable world model method that incorporates relevant global information, to train policies purely using synthetic trajectories of latent states generated by the model. The key insights behind MABL are that for each agent, **1.** relevant global information can be encoded into a global latent state to learn policies through centralized training, and **2.** the global latent state need not be used during execution and can instead be used to inform the representation learning of a separate agent-specific latent state used during decentralized execution. Specifically, as shown in Figure 1, MABL introduces a novel bi-level model to learn a hierarchical latent space. At the top level, the model learns the global latent state; at the bottom level, it learns an agent latent state, conditioned on the global latent state. MABL can be used as an additional module with any existing MARL algorithm. Our empirical studies on the challenging StarCraft Multi Agent Challenge (SMAC) [32], Flatland [26], and Multi-Agent MuJoCo (MAMuJoCo) [30] benchmarks show that MABL outperforms state-of-the-art multi-agent latent-variable world models in sample efficiency across a variety of discrete and continuous multi-agent tasks.

## 2 PRELIMINARIES

*Multi Agent Reinforcement Learning.* We consider MARL in a partially observable Markov game [27]. The game is represented as $G = \langle N, S, \mathbf{A}, P, R^i, \{O^i\}, \{O^i\}, \gamma \rangle$. $N = \{1, \ldots, n\}$ is the set of agents, $S$ the set of states, and $\mathbf{A} = \prod A^i$ the joint action space, where $A^i$ is the action space for agent $i$. At timestep $t$, agent $i \in N$ receives an observation $o_t^i$ governed by the observation function $O^i(s) : S \to O^i$, and chooses an action $a_t^i \in A^i$. Given the current state $s_t$ and the agents' joint action $\mathbf{a}_t = \{a_t^i\}_{i=1}^n$, the environment transitions to the next state $s_{t+1}$ according to the state transition function $P(s_{t+1}|s_t, \mathbf{a}_t) : S \times \mathbf{A} \times S \to [0, 1]$. Each agent then receives a reward $r_t^i$ according to its reward function $R^i : S \times a^i \to \mathbb{R}$. Each agent takes actions according to its policy $\pi^i(a_t^i|\tau_t^i)$, which is conditioned on its action-observation history $\tau_t^i$. Together, these policies comprise the joint policy $\boldsymbol{\pi}$, which induces the action-value function for each agent $i$, $Q_i^{\boldsymbol{\pi}} = \mathbb{E}_{\boldsymbol{\pi}}[\sum_{j=0}^{\infty} \gamma^j r_{t+j}^i]$, where $\gamma \in [0, 1]$ is the discount factor. In model-free MARL, agents do not know $P$ or $R$ and must learn policies that maximize $Q$ by interacting with the environment.

*Model-Based Reinforcement Learning.* In contrast to model-free methods, model-based RL methods learn an explicit model trained to estimate the environment dynamics (i.e., state transition $P$ and reward function(s) $R$) using self-supervised learning [14, 20, 37]. We consider the most popular style of model-based RL methods, which follow the Dyna algorithm [37], where the model of the environment dynamics, called a world model, is learned from real environment interactions. To deal with high-dimensional inputs, model-based RL algorithms have employed latent-variable world models [12, 13, 20] that learn and generate trajectories of compact, low-dimensional latent states $z$ from observations and actions, $o$ and $a$, as input during policy learning. In contrast, previous work generates synthetic data in the original high-dimensional space.

*Latent-Variable World Models.* Latent-variable world models are implemented as sequential variational auto-encoders [17] for learning environment dynamics. More concretely, consider a partially observable Markov decision process (POMDP) [3] described by $\langle S, A, P, R, O, O, \gamma \rangle$, where the symbols mean the same as before, except with a single agent. Given training data consisting of observation and action sequences $\{o_1, a_0, \ldots, o_T, a_{T-1}\}$, we can train the sequential variational auto-encoders with latent variables $z_t$ to maximize the probability of the data $p(o_{1:T}|a_{0:T-1})$. Directly maximizing this probability is challenging, so a typical approach is to consider the Evidence Lower Bound (ELBO) [17] for the log-likelihood of the sequence of observations:

$$\log p(o_{1:T}|a_{0:T-1}) \geq \mathbb{E}_{z_{1:T} \sim q} \sum_{t=1}^{T} \bigg[ \log p(o_t|z_t) \\ - D_{KL}\big(q(z_t|o_t, a_{t-1}) \| p(\hat{z}_t|a_{t-1})\big) \bigg],$$

where $D_{KL}$ refers to the KL divergence. The latent-variable model thus consists of a transition model representing the prior distribution $p(\hat{z}_t|a_{t-1})$, a representation model representing the posterior distribution $q(z_t|o_t, a_{t-1})$, and an observation decoder $p(\hat{o}_t|z_t)$ to reconstruct observations $\hat{o}_t$ from latent states $z_t$. All components are parameterized by neural networks and trained through amortized variational inference [17]. Once trained, $z_t$ serves as the compact latent state at time $t$, and the model can be used to generate synthetic trajectories of latent states for RL training.

## 3 MULTI-AGENT BI-LEVEL WORLD MODEL

We propose Multi-Agent Bi-Level world model (MABL), a novel model-based MARL algorithm that uses a latent-variable world model architecture. Crucially, our model leverages the insight that vital global information — such as the observations of all agents and any extra information available during training — does not itself need to be used during execution, as long as we use it to inform the representation that is used during execution. In this way, we achieve centralized training with decentralized policy execution.

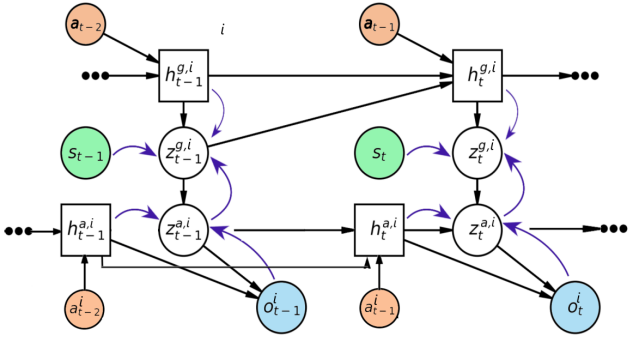

**Figure 2:** Transition dynamics components of the bi-level latent-variable model. Shaded circled nodes represent inputs, unshaded circled nodes represent random variables, and square nodes represent deterministic embeddings. The transition model and recurrent models are shown using black arrows; the representation model is shown using blue arrows.

Concretely we propose a bi-level latent-variable world model. The *global latent state* at the top level of the hierarchy encodes information about the global state of the environment relevant to the agent's learning. The global latent state is used to inform the learning of an *agent latent state* at the bottom level of the hierarchy to encode agent-specific local information. The agent latent state can be computed during execution time using just the agent's observation, meaning that it is more informative (because it encodes information derived from the global state during centralized training) but also more useful (because it can be computed and used in a decentralized manner during execution time).

In this section, we first describe our novel bi-level world model architecture and explain how it encodes relevant global information. We then detail the training framework for the model and MARL algorithm that is trained with the latent trajectories generated by the model. Throughout this section, we describe the algorithm with respect to an agent $i$.

## 3.1 Bi-Level World Model

To effectively capture environment dynamics in multi-agent settings, we introduce a novel bi-level architecture for the world model. The bi-level world model embeds high-dimensional inputs into latent representations to form a predictive model, as we now explain.

The model takes as input multi-agent trajectories of length $T$, represented as $\{o_t^i, s_t, \mathbf{a}_{t-1}, r_t^i\}_{t=1}^T$. The input trajectories are sampled from a buffer, which we call the model buffer, populated through interactions of the agents with the environment. Our model comprises of neural networks that serve two main functions: learning the transition dynamics and supporting the trajectory generation for MARL training. We refer to the former as *Transition Dynamics* components and the latter as *Auxiliary* components. All components are parameterized by neural networks with combined weights $\psi$ and are trained jointly. Each agent has a bi-level model, but the parameters ($\psi$) of model are shared among all agents to ensure scalability to settings with a large number of agents.

*3.1.1 **Transition Dynamics Components**.* We first describe the *Transition Dynamics* components, as illustrated in Figure 2. They are the recurrent models, the representation model, the transition model, and the observation model:

Recurrent Models: 
$$\begin{cases} \text{Global: } h_t^{g,i} = f_\psi^{g,i}(h_t^{g,i}|h_{t-1}^{g,i}, z_{t-1}^{g,i}, \mathbf{a}_{t-1}) \\ \text{Agent: } h_t^{a,i} = f_\psi^{a,i}(h_t^{a,i}|h_{t-1}^{a,i}, z_{t-1}^{a,i}, a_{t-1}) \end{cases}$$

Representation Model: 
(Posterior Distribution) 
$$\begin{cases} \text{Global: } z_t^{g,i} \sim q_\psi(z_t^{g,i}|s_t, z_t^{a,i}, h_t^{g,i}) \\ \text{Agent: } z_t^{a,i} \sim q_\psi(z_t^{a,i}|o_t^i, h_t^{a,i}) \end{cases}$$

Transition Model: 
(Prior Distribution) 
$$\begin{cases} \text{Global: } \hat{z}_t^{g,i} \sim p_\psi(\hat{z}_t^{g,i}|h_t^{g,i}) \\ \text{Agent: } \hat{z}_t^{a,i} \sim p_\psi(\hat{z}_t^{a,i}|h_t^{a,i}, \hat{z}_t^{g,i}) \end{cases}$$

Observation Model: $\quad \hat{o}_t^i = p_\psi(\hat{o}_t^i|h_t^{a,i}, z_t^{a,i}).$

*Recurrent Models.* To accurately learn multi-agent environment dynamics, the latent states should not only capture information about the current state of the environment but also past states and actions, especially in partially observable settings. The goal of the recurrent models is to capture this relevant historical information with deterministic embeddings. The global recurrent model propagates information about past environment states and joint actions through its embeddings $h_t^{g,i}$. In contrast, the agent recurrent model captures information about the action-observation history of the agent through $h_t^{a,i}$. Both recurrent models are implemented as Recurrent Neural Networks (RNNs) [24], and the embeddings are computed as hidden states of the RNNs.

*Representation Model.* The representation model learns the overall posterior distribution, which we factorise into global and agent posterior distributions over the global ($z_t^{g,i}$) and agent ($z_t^{a,i}$) latent states, as they are potentially easier to learn. We implement the posterior latent states as vectors of multiple categorical variables as in [13].

*Transition Model.* The role of the transition model is to predict future global and agent latent states without access to the environment global state and the agent observation that causes them. This way, the transition model can be used to generate synthetic trajectories for policy learning. Specifically, it learns the overall prior distribution over global ($\hat{z}_t^{g,i}$) and agent ($\hat{z}_t^{a,i}$) latent states, factorised into global and agent prior distributions, similar to the posterior. One key difference between the posterior latent states and the prior latent states is that $z_t^{g,i}$ is conditioned on $s_t$ and $z_t^{a,i}$ is conditioned on $o_t^i$. As with the representation model, the prior latent states are vectors of multiple categorical variables.

*Observation Model.* The observation model outputs a prediction of the current observation $\hat{o}_t^i$, given the agent embedding $h_t^{a,i}$ and the agent posterior latent state $z_t^{a,i}$. It is required to train the bi-level model using amortized variational inference.

*Benefits of Bi-Level Structure.* We now explain the benefits of having a bi-level structure. At each timestep $t$, the agent prior latent state $\hat{z}_t^{a,i}$ is conditioned on the global prior latent state $\hat{z}_t^{g,i}$ in a top-down fashion. Simultaneously, the global posterior latent state $z_t^{g,i}$ is conditioned on the agent prior latent state $z_t^{a,i}$ in a bottom-up fashion. This conditioning ensures a flow of information between the top and bottom levels of the latent-variable model, leading to a structured hierarchy of latent states. We design the representation model to enable inference of the agent posterior latent $z_t^{a,i}$ during execution without computing $z_t^{g,i}$. The latent-variable model thus incorporates relevant global information without violating the CTDE paradigm.

### 3.1.2 Auxiliary Components.

The goal of our model is to generate synthetic trajectories for MARL training. At a minimum, to learn how to act using synthetic trajectories, MARL agents additionally require feedback in terms of reward and knowledge of whether a state is terminal [12, 13]. Furthermore, in some environments, the availability of actions changes at each timestep [7]. As a result, we include auxiliary components in our model for predicting these values over trajectories.

The auxiliary components are implemented as neural networks, one each for predicting the reward, episode termination, and available actions at each timestep $t$. We predict rewards using a reward predictor network that outputs a continuous value $\hat{r}_t^i$. Empirically, we find that feeding just $z_t^{a,i}$ and $h_t^{a,i}$ to the reward predictor results in better overall performance. The termination predictor predicts whether the current state is terminal or not by outputting $\hat{\gamma}_t^i$, a binary value that is 1 if the episode terminates at time $t$. The available action predictor predicts $\hat{A}_t^{s,i}$, a vector of size $A$, each value of which denotes whether that action is available at time $t$. Both the termination and available action predictors are implemented as Bernoulli distributions, and take as input, $z_t^{a,i}$, $z_t^{g,i}$, $h_t^{a,i}$ and $h_t^{g,i}$.

Though not used to generate trajectories, our model also includes an action decoder neural network to reconstruct each agent's action $\hat{a}_t^i$ as an auxiliary component. As described in prior work [7, 16], the action decoder encourages the model to encode agent-specific information into the latent states. It does so by maximizing mutual information between each agent's latent state and its action. The auxiliary components are:

$$\text{Reward Predictor:} \quad \hat{r}_t^i \sim p_\psi(\hat{r}_t^i | z_t^{a,i}, h_t^{a,i})$$
$$\text{Termination Predictor:} \quad \hat{\gamma}_t^i \sim p_\psi(\hat{\gamma}_t^i | z_t^{a,i}, z_t^{g,i}, h_t^{a,i}, h_t^{g,i})$$
$$\text{Available Action Predictor:} \quad \hat{A}_t^{s,i} \sim p_\psi(\hat{A}_t^{s,i} | z_t^{a,i}, z_t^{g,i}, h_t^{a,i}, h_t^{g,i})$$
$$\text{Action Decoder:} \quad \hat{a}_t^i \sim p_\psi(\hat{a}_t^i | z_t^{a,i}, z_t^{g,i}, h_t^{a,i}, h_t^{g,i}).$$

## 3.2 Training the Model

Having described the bi-level model architecture, we now explain the loss function used to train the model. We train all components of our model jointly with the loss $\mathcal{L}(\psi)$. The loss is a sum of multiple terms. We write the total loss $\mathcal{L}(\psi)$ as:

$$\mathcal{L}(\psi) = \mathcal{L}_{\text{ELBO}} + \mathcal{L}_{\hat{r}_t} + \mathcal{L}_{\hat{\gamma}_t} + \mathcal{L}_{\hat{A}_t} + \mathcal{L}_{\hat{a}_t}.$$

The first term is the ELBO loss $\mathcal{L}_{\text{ELBO}}$, which trains the transition dynamics components to maximize the ELBO under the data generating distribution $p(o_{1:T}^i | \mathbf{a}_{0:T-1})$ using amortized variational inference.

We provide a detailed derivation of the ELBO in Appendix A. We write it as:

$$\mathcal{L}_{\text{ELBO}} = -\sum_{t=1}^{T} \log p_\psi(\hat{o}_t^i | z_t^{a,i}, h_t^{a,i})$$
$$+ D_{KL}\Big(q_\psi(z_t^{a,i} | o_t^i, h_t^{a,i}) \| p_\psi(\hat{z}_t^{a,i} | h_t^{a,i}, \hat{z}_t^{g,i})\Big)$$
$$+ D_{KL}\Big(q_\psi(z_t^{g,i} | s_t, z_t^{a,i}, h_t^{g,i}) \| p_\psi(\hat{z}_t^{g,i} | h_t^{g,i})\Big).$$

The first term in $\mathcal{L}_{\text{ELBO}}$ corresponds to maximizing the log likelihood of the observations, given $z_t^{a,i}$ and $h_t^{a,i}$. The second and third terms together minimize the KL divergence ($D_{KL}$) between the overall prior and posterior distributions, $p_\psi(.)$ and $q_\psi(.)$. We have two KL divergence terms as we factorize the overall prior and posterior distributions into global and agent distributions. To ensure that the distributions are learnt effectively, we use KL balancing [13].

The remaining terms in $\mathcal{L}(\psi)$ train the auxiliary components to maximize the log likelihoods of their corresponding targets, given the latent states from the representation model and the deterministic embeddings from the recurrent models. These are:

$$\text{Reward:} \quad \mathcal{L}_{\hat{r}_t} = -\sum_{t=1}^{T} \log p_\psi(\hat{r}_t^i | z_t^{a,i}, h_t^{a,i})$$

$$\text{Termination:} \quad \mathcal{L}_{\hat{\gamma}_t} = -\sum_{t=1}^{T} \log p_\psi(\hat{\gamma}_t^i | z_t^{a,i}, z_t^{g,i}, h_t^{a,i}, h_t^{g,i})$$

$$\text{Available action:} \quad \mathcal{L}_{\hat{A}_t} = -\sum_{t=1}^{T} \log p_\psi(\hat{A}_t^{s,i} | z_t^{a,i}, z_t^{g,i}, h_t^{a,i}, h_t^{g,i})$$

$$\text{Action decoder:} \quad \mathcal{L}_{\hat{a}_t} = -\sum_{t=1}^{T} \log p_\psi(\hat{a}_t^i | z_t^{a,i}, z_t^{g,i}, h_t^{a,i}, h_t^{g,i}).$$

## 3.3 Learning Multi-Agent Behavior

A benefit of our method is that model learning is independent of the MARL algorithm used for policy learning. This allows us to use any off-the-shelf value-based or actor-critic MARL algorithm. In this section, we explain how we can train a generic actor-critic MARL algorithm using latent trajectories generated by our model.

Each agent is equipped with a policy, or actor, $\pi_\theta^i$, which is implemented as a neural network with parameters $\theta$ and trained to maximize the MARL objective. At timestep $t$ of environment interaction, the agent receives as input its agent latent state ($z_t^{a,i}$ during execution and $\hat{z}_t^{a,i}$ during training) and outputs an action:

$$a_t^i \sim \pi_\theta(a_t^i | z_t^{a,i} / \hat{z}_t^{a,i}, h_t^{a,i}).$$

MABL infers $z_t^{a,i}$ solely from its current observation and its agent embedding ($h_t^{a,i}$), facilitating decentralized policy execution.

Each agent is also equipped with a critic $V_\phi^i$ that is represented by a neural network with parameters $\phi$ and outputs an estimate $\hat{V}$ of the value function:

$$\hat{V}_t^i \sim V_\phi^i(\hat{z}_t^{a,i}, h_t^{a,i}, h_t^{g,i}).$$

As the critic is *centralized*, its input is a concatenation of the global latent state and the agent latent state, which we represent by $\hat{z}_t^i$. As in [7], the critic includes a self-attention mechanism [39]. The actor

**Algorithm 1** MABL: Learning a Multi-Agent Bi-Level World Model

1: Initialize shared actor $\pi_\theta$, critic $V_\phi$, bi-level model $M_\psi$ and model buffer $\mathcal{D}$
2: **for** $n \in N$ episodes **do**  ▷ **Environment Interaction**
3:     Collect an episode of environment data using $\pi_\theta$
4:     Store data in $\mathcal{D}$
5:     **for** $m \in M$ model training steps **do**  ▷ **Model Training**
6:         Draw $\mathcal{B}_\mathcal{M}$ sequences uniformly from $\mathcal{D}$
7:         Train model $M_\psi$ on $\mathcal{B}_\mathcal{M}$ via loss $\mathcal{L}(\psi)$
8:     **end for**
9:     **for** $r \in R$ policy learning steps **do**  ▷ **MARL Training**
10:        Draw $\mathcal{B}_\mathcal{R}$ sequences uniformly from $\mathcal{D}$
11:        Generate latent trajectories $\mathcal{D}_\mathcal{L}$ from $\mathcal{B}_\mathcal{R}$ using $M_\psi$, $\pi_\theta$
12:        Compute MARL objective on $\mathcal{D}_\mathcal{L}$
13:        Update $\pi_\theta$ and $V_\phi$ using the MARL objective
14:     **end for**
15: **end for**

and critic network parameters are shared by all agents to facilitate faster training in tasks that involve large numbers of agents [46].

If we were to instead use a value-based MARL algorithm (e.g, Q-MIX [31]), for policy learning, each agent's critic can be represented by $Q_i(\hat{z}_t^{a,i}, h_t^{a,i})$ and the global critic can be represented by:

$$Q_{tot}(Q_1(\hat{z}_t^{a,1}, h_t^{a,1}), ..., Q_n(\hat{z}_t^{a,n}, h_t^{a,n}), \hat{z}_t^{g,1}, ..., \hat{z}_t^{g,n}).$$

We learn multi-agent behavior purely within the latent-variable model. By this, we mean that the trajectories used for training consist of latent states generated by the model. We now detail the iterative procedure [11, 33] outlined in Algorithm 1 that we use for training MABL.

First, the MARL agents interact with the environment to collect real environment data (Lines 3 and 4). These trajectories are stored in the model buffer $\mathcal{D}$ to use for model training. Second, the bi-level model is trained using trajectories sampled from $\mathcal{D}$ (Lines 5-8). We then freeze the weights of the bi-level model in preparation for MARL training.

Third, the MARL training occurs using synthetic trajectories of length $H$ generated by the model (Lines 9-14). Specifically, $\mathcal{B}_R$ sequences in the form of states, observations, and previous joint actions are drawn from $\mathcal{D}$ (Line 10). From each tuple of state, observations, and previous joint actions in a sequence, corresponding global and agent latent states are then computed by the representation model. For $H - 1$ timesteps that follow, at each timestep, $t$, the agents choose a joint action $\mathbf{a}_t$ according to their policies. The transition model and recurrent models then predict the next latent states $\hat{z}_{t+1}^{a,i}$ and $\hat{z}_{t+1}^{g,i}$. This process is repeated to generate trajectories of latent states of length $H$ starting from each tuple. Then, the auxiliary components of the model take as input the necessary components to predict rewards, termination conditions, and available actions to generate latent trajectories $\mathcal{D}_L$ of the form $\{z_1^{a,i}, z_1^{g,i}, \hat{z}_{2:H}^{a,i}, \hat{z}_{2:H}^{g,i}, h_{1:H}^{a,i}, h_{1:H}^{g,i}, r_{1:H}^i, \gamma_{1:H}^i, \hat{A}_{1:H}^i\}_{i=1}^n$.

Finally, the MARL algorithm is trained on $\mathcal{D}_L$ (Lines 12 and 13) following the CTDE paradigm. We choose the popular actor-critic MARL algorithm Multi-Agent PPO (MAPPO) [46] for policy learning, as it has achieved strong results in various multi-agent tasks.

## 4 EXPERIMENTAL EVALUATION

In this section, we present an empirical study of the sample efficiency of MABL against state-of-the-art algorithms on three challenging benchmarks: SMAC [32], Flatland [26], and MAMuJoCo [30]. First, we perform a comparative evaluation of MABL against other CTDE multi-agent latent-variable world models. On observing strong performance gains, we then ask whether MABL would perform similarly to or better than even CTCE multi-agent latent-variable world models whose agents have access to the observations of all other agents during execution. Since MABL is a CTDE method, it is at a natural disadvantage in such a comparison. We then perform ablation studies to examine the attributes of the bi-level model that lead to MABL's performance gains. In summary, our empirical analysis is aimed at answering the following questions:

**RQ1:** Does MABL lead to better sample efficiency compared to the state-of-the-art CTDE multi-agent latent-variable world models?

**RQ2:** Does MABL lead to comparable sample efficiency compared to the state-of-the-art CTCE multi-agent latent-variable world models?

**RQ3:** Is the bi-level latent-variable model responsible for the improved sample efficiency? If so, what features of the model lead to these improvements?

*Environments.* We briefly describe the three benchmarks we use (SMAC, Flatland, and MAMuJoCo) and provide a detailed description in the Appendix B.3. For the SMAC benchmark, we conduct experiments on two *Easy* maps (2s vs 1sc and 3s vs 4z), one *Hard* map (3s vs 5z), and two *Super Hard* maps (Corridor and 3s5z vs 3s6z). We defer comparison plots for the 2s vs 1sc to Appendix B.4, as it is the easiest task across all environments and serves as a sanity check.

The Flatland benchmark is a discrete action-space 2D grid environment that simulates train traffic on a railway network. Each agent controls a train and receives a positive reward on reaching its destination and penalties for colliding with other agents or being late. We conduct experiments with the 5 and 10 agent variants. Both SMAC and Flatland have discrete action spaces. We also evaluate our algorithm on MAMuJoCo [30], a continuous multi-agent robotic control benchmark where each agent controls a portion of the joints that together control the robot. We conduct experiments on the 2-agent Humanoid and 2-agent Humanoid Standup environments.

*Experimental Details.* Because we aim to investigate improvements in sample efficiency, we adopt the low data regime established in prior work [7, 12, 15, 42]. In our experiments, we train each algorithm across 3 independent runs with the same number of environment steps. We ensure that each algorithm that uses world models also generates the same number of synthetic samples for training. In our comparisons, we use the same MARL algorithm, MAPPO, for policy learning. We detail the hyperparameters, neural network architecture, and implementation specifics in Appendix B and make our code available here.

*Baselines.* We compare MABL against state-of-the-art CTDE and CTCE baselines. The CTDE baselines are Dreamer-v2 [13] and MAPPO [46]. Dreamer-v2 is a state-of-the-art single-agent model-based RL algorithm, which we implement as a multi-agent algorithm by learning a latent-variable world model independently for each agent. Including MAPPO trained solely on real environment data as a baseline serves to examine the effectiveness of our model-based baselines, all of which train MAPPO on synthetic latent trajectories. The CTCE baselines are MAMBA [7] and MAG [42], which are the current state-of-the-art in MARL with latent-variable world models. MAMBA [7] uses a transformer encoder block to aggregate observations of all agents to compute latent states for each agent, which are then input to agent policy networks. Because MAMBA requires the transformer block to compute latent states during execution, each agent requires access to the inputs of other agents during execution, resulting in centralized policy execution. MAG [42] is a recent work that builds upon MAMBA to achieve state-of-the-art sample efficiency on SMAC. MAG has one world model per agent, with a separate set of model parameters per agent. It uses the same model architecture as MAMBA, inheriting the CTCE property from it. The difference from MAMBA is that the world models are now trained jointly to interact with each other, taking into account the long-term joint effect of local predictions at each step to generate trajectories with lower cumulative errors [42]. We provide a conceptual comparison of all model-based MARL algorithms we consider in Appendix B.5.

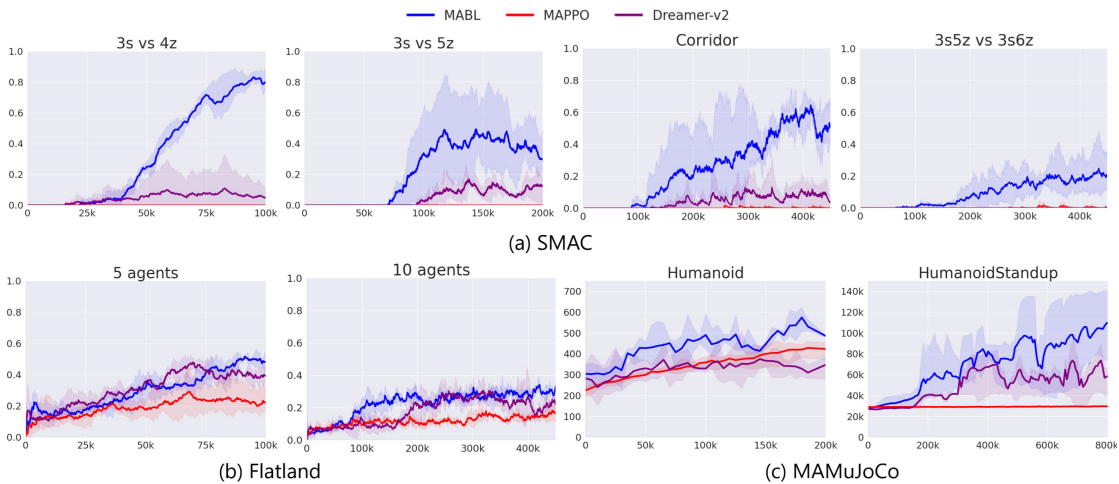

Figure 3: Comparisons against CTDE baselines across all environments. Curves represent the mean over 3 independent runs, and shaded regions show the minimum and maximum scores. X axis shows the number of steps taken in the real environment; Y axis denotes the win-rate for SMAC, and the reward for Flatland and MAMuJoCo. Plots are smoothed using exponential moving average. MABL not only achieves better overall performance than the baselines, but does so more rapidly, underscoring its sample efficiency.

### 4.1 Performance Comparison: CTDE methods

We now compare MABL with CTDE baselines. In Figure 3, we present the performance of MABL and CTDE baselines on the entirety of the low data regime. In Table 1, we summarize the final performance of all algorithms over the last 15k environment steps. These results show that, except on the *Easy* 2s vs 1sc SMAC map, MABL consistently outperforms all CTDE baselines in terms of sample efficiency by large margins, answering **RQ1**. On the *Easy* 2s vs 1sc map, the win-rate of MABL is close to the best-performing baseline MAPPO while Dreamer-v2 fails to learn a good policy. In addition, MABL initially achieves a high win-rate of 80% nearly **2x** faster than MAPPO on this task (Appendix B.4). The performance gains of MABL are significant in challenging SMAC environments such as Corridor and 3s5z vs 3s6z, as well as HumanoidStandup, demonstrating MABL's capability to handle complex tasks. The superior overall score and sample efficiency of MABL reveals the benefit of incorporating global information into the representation learning of latent-variable world models.

### 4.2 Performance Comparison: CTCE Methods

Given the surprisingly strong performance of MABL compared with CTDE baselines, we now compare MABL against state-of-the-art CTCE methods. CTCE methods access global information at execution time, putting our method at a disadvantage. Should MABL perform comparably or better than CTCE methods, it would suggest that the bi-level model captures more pertinent information in its latent states for policy learning than existing state-of-the-art techniques.

We plot the performance of MABL compared to CTCE baselines in Figure 4. Surprisingly, from Table 1 and Figure 4, we observe that MABL achieves superior or comparable sample efficiency to both MAMBA and MAG on all benchmarks. MABL outperforms MAMBA on all environments except on the *Easy* 2s vs 1sc SMAC map and the *Super Hard* 3s5z vs 3s6z map. MABL and MAMBA achieve nearly a 100% win-rate on 2s vs 1sc. On 3s5z vs 3s6z, both algorithms achieve a win-rate of nearly 20% after just 450k environment steps.

We see a similar trend in performance when comparing MABL and MAG. MABL outperforms MAG on Flatland, MAMuJoCo, and all SMAC maps — except 2s vs 1sc, where both algorithms achieve nearly perfect performance, and the *Super Hard* Corridor map, where they exhibit similar levels of performance. Overall, MABL either matches or outperforms even state-of-the-art CTCE methods, answering **RQ2** in the affirmative.

| Benchmark | Map/Environment | Steps | MABL | CTDE Dreamer-v2 | MAPPO | CTCE MAMBA | MAG |
|---|---|---|---|---|---|---|---|
| SMAC | 2s vs 1sc | 100k | 92±7 | 37±40 | **98±3** | 94±7 | 87±23 |
| | 3s vs 4z | 100k | **83±18** | 4±11 | 0 | 24±23 | 30±32 |
| | 3s vs 5z | 200k | **31±30** | 14±24 | 0 | 2±8 | 5±13 |
| | Corridor | 450k | **52±31** | 5±16 | 0 | 31±30 | 55±27 |
| | 3s5z vs 3s6z | 450k | **19±17** | 0 | 0 | 19±27 | 16±23 |
| Flatland | 5 agents | 100k | **51±27** | 40±26 | 22±22 | 29±27 | 19±19 |
| | 10 agents | 450k | **31±19** | 22±20 | 18±16 | 29±27 | 13±14 |
| MAMuJoCo | Humanoid | 200k | **550±90** | 304±23 | 429±36 | 328±27 | 423±104 |
| | HumanoidStandup | 800k | **106k±29k** | 54k±11k | 30k±0.3k | 83k±30k | 76k±22k |

Table 1: Comparison of the average win-rate (% for SMAC)/reward, and standard deviation in win-rate/reward over the last 15k environment steps across environments. Numbers in bold indicate the highest mean performance among all CTDE methods. Except on the *Easy* 2s vs 1sc map, MABL outperforms CTDE baselines on all environments. MABL also either outperforms or performs similarly to the best CTCE baseline on all tasks.

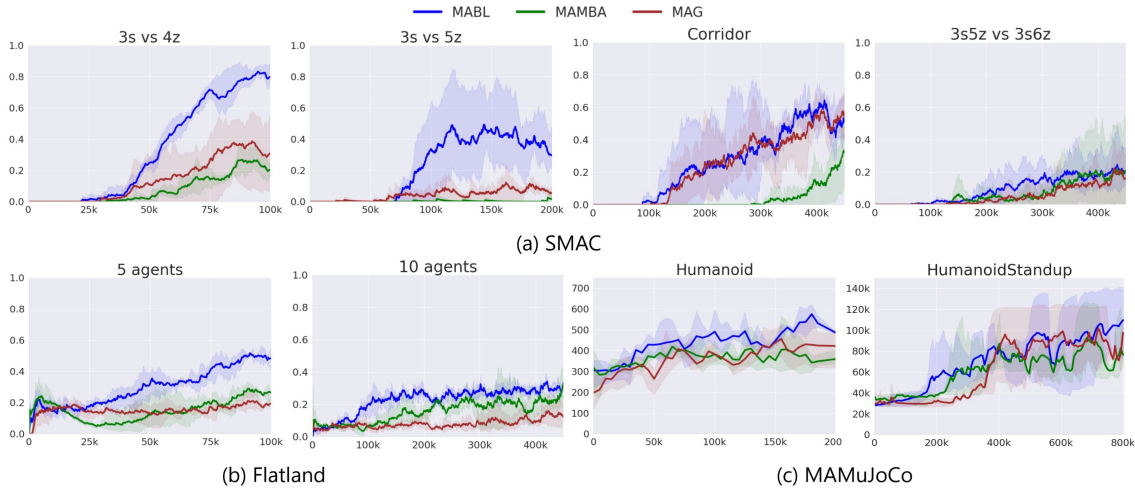

Figure 4: Comparisons against CTCE baselines across all environments. Curves represent the mean over 3 independent runs, and shaded regions show the minimum and maximum scores. X axis shows the number of steps taken in the real environment; Y axis denotes the win-rate for SMAC and the reward for Flatland and MAMuJoCo. Plots are smoothed using exponential moving average. MABL either matches or outperforms the CTCE baselines across all environments.

A surprising observation is that the CTDE baselines perform better overall than the CTCE baselines on the 5-agent Flatland task. This result suggests that, in relatively less complex environments where each agent's observation has enough information to learn good behavior, simpler algorithms like Dreamer-v2 may excel in representation learning compared to MAMBA or MAG. The results also show that CTCE baselines perform more effective representation learning than the CTDE baselines on more complex environments. However, MABL demonstrates consistent high performance across all environments, pointing to improved representation learning compared to both CTDE and CTCE baselines. In the next section, we investigate this further through an ablation study.

### 4.3 Ablation Study

Given the performance of MABL compared to both CTDE and CTCE methods, we now seek to understand the attributes of MABL model that contribute most to these gains in sample efficiency. Suspecting that better representation learning using the bi-level model is responsible for these gains, we ablate two attributes of our model.

First, we ablate the *global latent state*. Our model learns a bi-level latent space and is capable of encoding global information into the upper-level global latent state. To understand the importance of the global latent state for policy learning, we remove the upper level. The model we obtain is the same as Dreamer-v2 and learns only the agent latent state with access to the observation of the agent. In our plots, we refer to this variant as Dreamer-v2.

Second, we suspect that *learning one global latent state per agent* enables better representation learning, as the model can more readily incorporate information relevant to each agent into its global latent state. To test this, we modify the upper level of the model such that we learn a single shared global latent state instead of $N$ separate ones. Because each of the $N$ lower-level agent latent states is conditioned on the single global latent, the model cannot encode relevant global information for each agent into the agent's respective global latent. We call the resulting variant MABL-SG (Single Global).

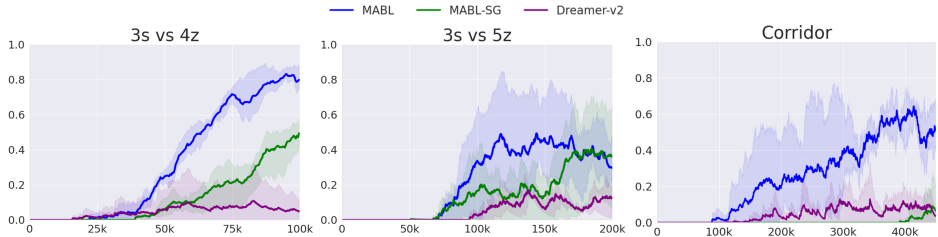

Figure 5: Training curves of ablation studies. X axis denotes the number of environment steps. Plots are smoothed using an exponential moving average. Shaded regions show the maximum and minimum win-rate. MABL outperforms all ablations, implicating the crucial role of the bi-level model.

We perform ablation studies on one SMAC map from each level of difficulty: 3s vs 4z (*Easy*), 3s vs 5z (*Hard*), and Corridor (*Super Hard*). We visualize the training curves in Figure 5. On all maps, MABL achieves greater sample efficiency than the two variants. Because Dreamer-v2 exhibits the lowest performance overall, we believe that the bi-level model's ability to incorporate global information through the global latent state is most responsible for MABL's gains in sample efficiency. The decreased performance of MABL-SG compared to MABL supports our hypothesis that learning per-agent global latents, as opposed to a single shared global latent, improves policy learning. We suspect that the use of a single, shared global latent facilitates the encoding of noisy and irrelevant information, which impedes performance. Taken together, the superior performance of MABL highlights the crucial role of the bi-level model. In particular, these experiments suggest that the bi-level model successfully incorporates relevant global information into the global latent and learns a structured hierarchy of latent states, answering **RQ3**.

## 5 RELATED WORK

The majority of work in MARL focuses on the model-free setting [8, 22, 28]. Despite their impressive performance, model-free MARL algorithms often suffer from a high sample complexity. Several approaches have been developed to address this issue. One line of work [31, 34, 36, 45] uses the insight of value decomposition [5]: value functions can be decomposed into simpler functions which can be learned more easily and (relatively) independently. They can then be recombined to approximate the original, more complex value function. Another line of work focuses on actor-critic methods [8, 22, 30, 46] that learn a centralized critic conditioned on global state and joint action to reduce non-stationarity and improve sample efficiency. Because we can use any model-free MARL algorithm for learning multi-agent policies in latent space, these techniques are complementary to our contributions.

In MARL, latent-variable models have shown promise in learning the reward function in inverse RL [10] and representations of competing agents' strategies [43]. We focus on latent-variable world models in MARL. While prior work [19] uses a multi-step latent-variable world model in 2-player games to predict future joint observations and actions, it is only applicable in 2-agent scenarios, does not generate synthetic data, and does not follow the CTDE paradigm. Only recently have latent-variable world models been used to learn environment dynamics in Markov games to improve sample efficiency [7, 42]. These approaches are based on direct extensions of single-agent methods that use latent-variable world models [13]. The dynamics of each agent is learned as though the agent is in a POMDP, and deep learning architectures, specifically, the transformer [21, 39], aggregate latent states from all agents to reduce non-stationarity and model errors [7, 42].

However, latent-variable models designed this way cannot be used for decentralized execution, as they require access to all observations to predict latent states. They also cannot incorporate any additional global information available during training In multi-agent tasks, encoding global information, such as the global state in SMAC, for example, is crucial to learn successful behaviors from latent states.

## 6 CONCLUSION

We presented a novel model-based MARL algorithm, MABL, that learns policies purely using latent trajectories generated by a bi-level latent-variable world model. Our model effectively learns environment dynamics in multi-agent tasks by factorizing the latent space into a high-level global latent state and a low-level agent latent state. We evaluated MABL across a variety of tasks in SMAC, Flatland and MAMuJoCo. MABL, which is a CTDE method, greatly outperforms state-of-the-art CTDE baselines in sample efficiency on all environments except for the simplest SMAC map, 2s vs 1sc. MABL either outperforms or performs similarly to even state-of-the-art CTCE baselines in sample efficiency across all environments. While we achieve gains in sample efficiency, the learned latent states are not interpretable. To deploy our method in a real-world scenario, future work should involve improving representation learning to achieve interpretability of the latent space.

## 7 ACKNOWLEDGEMENTS

This research was supported in part by NSF IIS-2046640 (CAREER). We thank NVIDIA for providing computing resources. We thank Robert Bosch Center for Data Science and AI for supporting author Aravind Venugopal's Post-Baccalaureate Fellowship for part of the duration of this work. We thank Rex Chen for his contributions towards setting up the computational resources for the experiments.

# REFERENCES

[1] Adrian K Agogino and Kagan Tumer. 2012. A multiagent approach to managing air traffic flow. *Autonomous Agents and Multi-Agent Systems* 24, 1 (2012), 1–25.

[2] J Andrew Bagnell and Jeff G Schneider. 2001. Autonomous helicopter control using reinforcement learning policy search methods. In *Proceedings 2001 ICRA. IEEE International Conference on Robotics and Automation (Cat. No. 01CH37164)*, Vol. 2. IEEE, 1615–1620.

[3] Anthony R Cassandra, Leslie Pack Kaelbling, and Michael L Littman. 1994. Acting optimally in partially observable stochastic domains. In *Aaai*, Vol. 94. 1023–1028.

[4] Dane Corneil, Wulfram Gerstner, and Johanni Brea. 2018. Efficient model-based deep reinforcement learning with variational state tabulation. In *International Conference on Machine Learning*. PMLR, 1049–1058.

[5] Thomas G Dietterich. 2000. Hierarchical reinforcement learning with the MAXQ value function decomposition. *Journal of artificial intelligence research* 13 (2000), 227–303.

[6] Joris Dinneweth, Abderrahmane Boubezoul, René Mandiau, and Stéphane Espié. 2022. Multi-agent reinforcement learning for autonomous vehicles: A survey. *Autonomous Intelligent Systems* 2, 1 (2022), 27.

[7] Vladimir Egorov and Aleksei Shpilman. 2022. Scalable Multi-Agent Model-Based Reinforcement Learning. *arXiv preprint arXiv:2205.15023* (2022).

[8] Jakob Foerster, Gregory Farquhar, Triantafyllos Afouras, Nantas Nardelli, and Shimon Whiteson. 2018. Counterfactual multi-agent policy gradients. In *Proceedings of the AAAI Conference on Artificial Intelligence*, Vol. 32.

[9] Sven Gronauer and Klaus Diepold. 2022. Multi-agent deep reinforcement learning: a survey. *Artificial Intelligence Review* (2022), 1–49.

[10] Nate Gruver, Jiaming Song, Mykel J Kochenderfer, and Stefano Ermon. 2020. Multi-agent adversarial inverse reinforcement learning with latent variables. In *Proceedings of the 19th International Conference on Autonomous Agents and MultiAgent Systems*. 1855–1857.

[11] David Ha and Jürgen Schmidhuber. 2018. World models. *arXiv preprint arXiv:1803.10122* (2018).

[12] Danijar Hafner, Timothy Lillicrap, Jimmy Ba, and Mohammad Norouzi. 2019. Dream to Control: Learning Behaviors by Latent Imagination. In *International Conference on Learning Representations*.

[13] Danijar Hafner, Timothy Lillicrap, Mohammad Norouzi, and Jimmy Ba. 2020. Mastering atari with discrete world models. *arXiv preprint arXiv:2010.02193* (2020).

[14] Michael Janner, Justin Fu, Marvin Zhang, and Sergey Levine. 2019. When to Trust Your Model: Model-Based Policy Optimization. *Advances in Neural Information Processing Systems* 32 (2019), 12519–12530.

[15] Lukasz Kaiser, Mohammad Babaeizadeh, Piotr Milos, Blazej Osinski, Roy H Campbell, Konrad Czechowski, Dumitru Erhan, Chelsea Finn, Piotr Kozakowski, Sergey Levine, et al. 2019. Model-based reinforcement learning for atari. *arXiv preprint arXiv:1903.00374* (2019).

[16] Nan Rosemary Ke, Amanpreet Singh, Ahmed Touati, Anirudh Goyal, Yoshua Bengio, Devi Parikh, and Dhruv Batra. 2019. Learning dynamics model in reinforcement learning by incorporating the long term future. *arXiv preprint arXiv:1903.01599* (2019).

[17] Diederik P Kingma and Max Welling. 2013. Auto-encoding variational bayes. *arXiv preprint arXiv:1312.6114* (2013).

[18] Maryam Kouzehgar, Malika Meghjani, and Roland Bouffanais. 2020. Multi-agent reinforcement learning for dynamic ocean monitoring by a swarm of buoys. In *Global Oceans 2020: Singapore–US Gulf Coast*. IEEE, 1–8.

[19] Orr Krupnik, Igor Mordatch, and Aviv Tamar. 2020. Multi-agent reinforcement learning with multi-step generative models. In *Conference on Robot Learning*. PMLR, 776–790.

[20] Alex Lee, Anusha Nagabandi, Pieter Abbeel, and Sergey Levine. 2020. Stochastic Latent Actor-Critic: Deep Reinforcement Learning with a Latent Variable Model. *Advances in Neural Information Processing Systems* 33 (2020).

[21] Tianyang Lin, Yuxin Wang, Xiangyang Liu, and Xipeng Qiu. 2022. A survey of transformers. *AI Open* (2022).

[22] Ryan Lowe, Yi Wu, Aviv Tamar, Jean Harb, Pieter Abbeel, and Igor Mordatch. 2017. Multi-agent actor-critic for mixed cooperative-competitive environments. *arXiv preprint arXiv:1706.02275* (2017).

[23] Laëtitia Matignon, Laurent Jeanpierre, and Abdel-Illah Mouaddib. 2012. Coordinated multi-robot exploration under communication constraints using decentralized markov decision processes. In *Twenty-sixth AAAI conference on artificial intelligence*.

[24] Larry Medsker and Lakhmi C Jain. 1999. *Recurrent neural networks: design and applications*. CRC press.

[25] Thomas M Moerland, Joost Broekens, Aske Plaat, Catholijn M Jonker, et al. 2023. Model-based reinforcement learning: A survey. *Foundations and Trends® in Machine Learning* 16, 1 (2023), 1–118.

[26] Sharada Mohanty, Erik Nygren, Florian Laurent, Manuel Schneider, Christian Scheller, Nilabha Bhattacharya, Jeremy Watson, Adrian Egli, Christian Eichenberger, Christian Baumberger, et al. 2020. Flatland-RL: Multi-agent reinforcement learning on trains. *arXiv preprint arXiv:2012.05893* (2020).

[27] George E Monahan. 1982. State of the art—a survey of partially observable Markov decision processes: theory, models, and algorithms. *Management science* 28, 1 (1982), 1–16.

[28] Kamal K Ndousse, Douglas Eck, Sergey Levine, and Natasha Jaques. 2021. Emergent social learning via multi-agent reinforcement learning. In *International Conference on Machine Learning*. PMLR, 7991–8004.

[29] Georgios Papoudakis, Filippos Christianos, Arrasy Rahman, and Stefano V Albrecht. 2019. Dealing with non-stationarity in multi-agent deep reinforcement learning. *arXiv preprint arXiv:1906.04737* (2019).

[30] Bei Peng, Tabish Rashid, Christian A Schroeder de Witt, Pierre-Alexandre Kamienny, Philip HS Torr, Wendelin Böhmer, and Shimon Whiteson. 2020. FACMAC: Factored Multi-Agent Centralised Policy Gradients. *arXiv preprint arXiv:2003.06709* (2020).

[31] Tabish Rashid, Mikayel Samvelyan, Christian Schroeder, Gregory Farquhar, Jakob Foerster, and Shimon Whiteson. 2018. Qmix: Monotonic value function factorisation for deep multi-agent reinforcement learning. In *International Conference on Machine Learning*. PMLR, 4295–4304.

[32] Mikayel Samvelyan, Tabish Rashid, Christian Schroeder De Witt, Gregory Farquhar, Nantas Nardelli, Tim GJ Rudner, Chia-Man Hung, Philip HS Torr, Jakob Foerster, and Shimon Whiteson. 2019. The starcraft multi-agent challenge. *arXiv preprint arXiv:1902.04043* (2019).

[33] Jürgen Schmidhuber. 1991. Curious model-building control systems. In *Proc. international joint conference on neural networks*. 1458–1463.

[34] Kyunghwan Son, Daewoo Kim, Wan Ju Kang, David Earl Hostallero, and Yung Yi. 2019. Qtran: Learning to factorize with transformation for cooperative multi-agent reinforcement learning. In *International conference on machine learning*. PMLR, 5887–5896.

[35] Wen Sun, Nan Jiang, Akshay Krishnamurthy, Alekh Agarwal, and John Langford. 2019. Model-based rl in contextual decision processes: Pac bounds and exponential improvements over model-free approaches. In *Conference on learning theory*. PMLR, 2898–2933.

[36] Peter Sunehag, Guy Lever, Audrunas Gruslys, Wojciech Marian Czarnecki, Vinicius Zambaldi, Max Jaderberg, Marc Lanctot, Nicolas Sonnerat, Joel Z Leibo, Karl Tuyls, et al. 2017. Value-decomposition networks for cooperative multi-agent learning. *arXiv preprint arXiv:1706.05296* (2017).

[37] Richard S Sutton. 1991. Dyna, an integrated architecture for learning, planning, and reacting. *ACM Sigart Bulletin* 2, 4 (1991), 160–163.

[38] Emanuel Todorov, Tom Erez, and Yuval Tassa. 2012. Mujoco: A physics engine for model-based control. In *2012 IEEE/RSJ international conference on intelligent robots and systems*. IEEE, 5026–5033.

[39] Ashish Vaswani, Noam Shazeer, Niki Parmar, Jakob Uszkoreit, Llion Jones, Aidan N Gomez, Łukasz Kaiser, and Illia Polosukhin. 2017. Attention is all you need. *Advances in neural information processing systems* 30 (2017).

[40] Tingwu Wang, Xuchan Bao, Ignasi Clavera, Jerrick Hoang, Yeming Wen, Eric Langlois, Shunshi Zhang, Guodong Zhang, Pieter Abbeel, and Jimmy Ba. 2019. Benchmarking model-based reinforcement learning. *arXiv preprint arXiv:1907.02057* (2019).

[41] Xihuai Wang, Zhicheng Zhang, and Weinan Zhang. 2022. Model-based Multi-agent Reinforcement Learning: Recent Progress and Prospects. *arXiv preprint arXiv:2203.10603* (2022).

[42] Zifan Wu, Chao Yu, Chen Chen, Jianye Hao, and Hankz Hankui Zhuo. 2023. Models as Agents: Optimizing Multi-Step Predictions of Interactive Local Models in Model-Based Multi-Agent Reinforcement Learning. *arXiv preprint arXiv:2303.17984* (2023).

[43] Annie Xie, Dylan Losey, Ryan Tolsma, Chelsea Finn, and Dorsa Sadigh. 2020. Learning Latent Representations to Influence Multi-Agent Interaction. In *Conference on Robot Learning*.

[44] Xu Xu, Youwei Jia, Yan Xu, Zhao Xu, Songjian Chai, and Chun Sing Lai. 2020. A multi-agent reinforcement learning-based data-driven method for home energy management. *IEEE Transactions on Smart Grid* 11, 4 (2020), 3201–3211.

[45] Yaodong Yang, Jianye Hao, Ben Liao, Kun Shao, Guangyong Chen, Wulong Liu, and Hongyao Tang. 2020. Qatten: A general framework for cooperative multiagent reinforcement learning. *arXiv preprint arXiv:2002.03939* (2020).

[46] Chao Yu, Akash Velu, Eugene Vinitsky, Yu Wang, Alexandre Bayen, and Yi Wu. 2021. The surprising effectiveness of ppo in cooperative, multi-agent games. *arXiv preprint arXiv:2103.01955* (2021).

## A DERIVATION OF ELBO

In this section, we discuss how we obtain the ELBO for training the latent-variable model. For a trajectory length of $T$, we write the joint distribution of data as:

$$p(o^i_{1:T}, z^{a,i}_{1:T}, z^{g,i}_{1:T}|\mathbf{a}_{0:T-1}) = \prod_{t=1}^{T} p_\psi(o^i_t|z^{a,i}_t, h^{a,i}_t) p_\psi(z^{a,i}_t|h^{a,i}_t, z^{g,i}_t) f^{a,i}_\psi(h^{a,i}_t|h^{a,i}_{t-1}, z^{a,i}_{t-1}, a^i_{t-1}) p_\psi(z^{g,i}_t|h^{g,i}_t) f^{g,i}_\psi(h^{g,i}_t|h^{g,i}_{t-1}, z^{g,i}_{t-1}, \mathbf{a}_{t-1}) \quad (1)$$

We approximate our posterior distribution as:

$$q(z^{a,i}_{1:T}, z^{g,i}_{1:T}|o^i_{1:T}, s_{1:T}, \mathbf{a}_{0:T-1}) = \prod_{t=1}^{T} q_\psi(z^{a,i}_t|o^i_t, h^{a,i}_t) f^{a,i}_\psi(h^{a,i}_t|h^{a,i}_{t-1}, z^{a,i}_{t-1}, a^i_{t-1}) q_\psi(z^{g,i}_t|s_t, z^{a,i}_t, h^{g,i}_t) f^{g,i}_\psi(h^{g,i}_t|h^{g,i}_{t-1}, z^{g,i}_{t-1}, \mathbf{a}_{t-1}) \quad (2)$$

Using Equations (1) and (2) and applying Jensen's inequality, we obtain the ELBO as follows:

$$\log p(o^i_{1:T}|\mathbf{a}_{0:T-1}) = \log \int_{z^{a,i}_{1:T}, z^{g,i}_{1:T}} p(o^i_{1:T}, z^{a,i}_{1:T}, z^{g,i}_{1:T}|\mathbf{a}_{0:T-1}) dz^{a,i}_{1:T} dz^{g,i}_{1:T}$$

$$\geq \mathbb{E}_{z^{a,i}_{1:T}, z^{g,i}_{1:T} \sim q}[\log p(o^i_{1:T}, z^{a,i}_{1:T}, z^{g,i}_{1:T}|\mathbf{a}_{0:T-1}) - \log q(z^{a,i}_{1:T}, z^{g,i}_{1:T}|o_{1:T}, s_{1:T}, \mathbf{a}_{0:T-1})]$$

$$= \mathbb{E}_{z^{a,i}_{1:T}, z^{g,i}_{1:T} \sim q}[\sum_{t=1}^{T} \log p_\psi(o^i_t|z^{a,i}_t, h^{a,i}_t) - D_{KL}\Big(q_\psi(z^{a,i}_t|o^i_t, h^{a,i}_t) || p_\psi(z^{a,i}_t|h^{a,i}_t, z^{g,i}_t)\Big) - D_{KL}\Big(q_\psi(z^{g,i}_t|s_t, z^{a,i}_t, h^{g,i}_t) || p_\psi(z^{g,i}_t|h^{g,i}_t)\Big)].$$

$$(3)$$

## B EXPERIMENTAL DETAILS

Here, we provide the implementation details, including architectural and hyperparameter choices. We also describe the environments in more detail. To implement SMAC and MAMuJoCo experiments, we use a system with AMD EPYC 7453 CPUs and an NVIDIA RTX A6000 GPU. Flatland experiments were run on a system with Intel(R) Xeon(R) Platinum 8268 CPUs and an Nvidia A-100 GPU.

### B.1 Hyperparameters

*Latent-Variable Model.* We implemented all components of the model as full connected neural networks with ReLU activations. The specific hyperparameters used for SMAC, Flatland, and MAMuJoCo are listed in Table 2.

| Hyperparameter | Flatland | SMAC | MAMuJoCo |
| --- | --- | --- | --- |
| Number of epochs | 40 | 60 | 60 |
| Number of sampled rollouts | 32 | 40 | 40 |
| Sequence length | 50 | 20 | 20 |
| Rollout Horizon | 15 | 15 | 15 |
| Buffer Size | 5e5 | 2.5e5 | 2.5e5 |
| Number of categoricals (local & global) | 32 | 32 | 32 |
| Number of classes (local & global) | 32 | 32 | 32 |
| KL balancing entropy weight | 0.2 | 0.2 | 0.2 |
| KL balancing cross entropy weight | 0.8 | 0.8 | 0.8 |
| Trajectories between updates | 1 | 1 | 1 |
| Hidden layer size | 400 | 256 | 256 |
| Number of hidden layers | 2 | 2 | 2 |
| Gradient clipping norm | 100 | 100 | 100 |

Table 2: Hyperparameters for training the model

*MARL algorithm.* We implemented the shared actor and critic as fully connected neural networks with ReLU activations. For the Flatland environment, we observed that feeding just the global latent state as input to the critic network yielded better results, as opposed to a concatenation of global and agent latent states. The hyperparameters used for SMAC, Flatland and MAMuJoCo are listed in Table 6.

*Learning Rates.* We report learning rates for the actor, critic and model, for each task in Table 4.

| Hyperparameter | Flatland | SMAC | MA-Mujoco |
|---|---|---|---|
| Batch size | 2000 | 2000 | 2000 |
| GAE $\lambda$ | 0.95 | 0.95 | 0.95 |
| Entropy coefficient | 0.001 | 0.001 | 0.001 |
| Entropy annealing | 0.99998 | 0.99998 | 0.99998 |
| Number of updates | 4 | 4 | 4 |
| Epochs per update | 5 | 5 | 5 |
| Gradient clipping norm | 100 | 100 | 100 |
| Discount factor $\gamma$ | 0.99 | 0.99 | 0.99 |
| Trajectories between updates | 1 | 1 | 1 |
| Hidden size | 400 | 256 | 256 |

Table 3: Hyperparameters for training the MARL algorithm

| Environment | Actor | Critic | Model |
|---|---|---|---|
| Flatland: 5 agents | 5e-4 | 5e-4 | 2e-4 |
| Flatland: 10 agents | 5e-4 | 5e-4 | 2e-4 |
| MAMuJoCo: Humanoid | 3e-4 | 1e-5 | 1e-5 |
| MAMuJoCo: HumanoidStandup | 3e-4 | 0.5e-5 | 0.5e-5 |
| SMAC: 2s vs 1sc | 1e-4 | 1e-3 | 3e-4 |
| SMAC: 3s vs 4z | 1e-4 | 1e-3 | 3e-4 |
| SMAC: 3s vs 5z | 1e-4 | 1e-3 | 5e-4 |
| SMAC: Corridor | 1e-4 | 1e-3 | 3e-4 |
| SMAC: 3s5z vs 3s6z | 1e-4 | 1e-3 | 3e-4 |

Table 4: Learning rates for all benchmarks and all environments

## B.2 Implementation Details

In our SMAC experiments, we use the official code implementation of MAMBA, MAG and the official implementation of MAPPO for the model-free version of MAPPO. In our Flatland experiments, we use the official MAMBA code and re-implemented MAPPO to integrate it with the MAMBA codebase which supports the Flatland environment. We based our code on the official MAMBA implementation for fair and convenient comparison.

## B.3 Environments

*SMAC.* We use the default environment settings provided by the authors. We also describe the maps we conduct experiments on:

Easy Maps: We use the 2s vs 1sc map and the 3s vs 4z map. In 2s vs 1sc, the ally units consist of 2 Stalkers who must team up against a single enemy unit: a Spine Crawler. In 3s vs 4z, the ally units consist of 3 Stalkers, battling against 4 Zealots.

Hard Maps: We use the 3s vs 5z map in which 3 Stalkers face 5 Zealots. To gain a victory, the 3 allied Stalkers have to make enemy units give chase, while at the same time ensuring that they keep enough distance from them to not incur critical damage. This strategy is called kiting [32] and requires precise control and team coordination.

Super Hard Maps: We use the Corridor and 3s5z vs 3s6z maps. In the corridor map, 6 allied Zealots face 24 enemy Zerglings. To win, agents have to make use of the terrain features of the map, collectively blocking a narrow region of the map to block enemy attacks from all directions. The 3s5z vs 3s6z map is an assymteric scenario with different kinds of allied units, which makes learning successful team behavior extremely difficult.

*Flatland.* We use the classic [7] environment settings for Flatland, as shown in Table 5.

*MAMuJoCo.* MAMuJoCo is a multi-agent version of the popular continuous single-agent robotic control benchmark MuJoCo [38]. A multi-agent system for a MAMuJoCo task is created by representing the given robot as a body graph where joints, which are represented by vertices are connected to each other by body segments, represented by edges. The body graph is then split into disjoint subgraphs. Each subgraph represents an agent and contains one or more controllable joints. An agent's action space is given by the joint vector of all actuators that can be controlled by it to move the joints. Each agent observes the positions of its body parts, and in our experiments, cannot observe the positions of the body parts of other agents. We conduct experiments on the Humanoid and Humanoid Standup tasks [30]. The Humanoid task consists of two agents, one controlling the upper body joints and one controlling the lower body joints below the torse of a Humanoid robot. Humanoid Standup consists of two agents, each of which controls the joints of one leg of a humanoid robot. In both tasks, the goal is to maximize +ve speed in the x direction. We limit episode length to a maximum of 1000 timesteps, for both tasks. We also use the default environment settings used in [30] and provided in the MAMuJoCo Github repository.

| Parameter | 5 agents | 10 agents |
|---|---|---|
| Height | 35 | 35 |
| Width | 35 | 35 |
| Number of cities | 3 | 4 |
| Grid distribution of cities | False | False |
| Maximum number of rails between cities | 2 | 2 |
| Maximum number of rails in cities | 4 | 4 |
| Malfunction rate | 1/100 | 1/150 |

Table 5: Flatland environment parameters

### B.4 2s vs 1sc Results

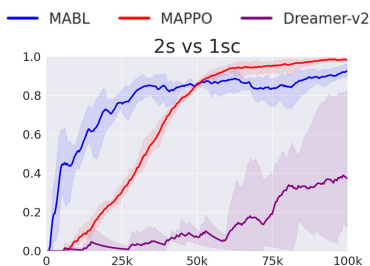

Figure 6: Comparisons against CTDE baselines on 2s vs 1sc

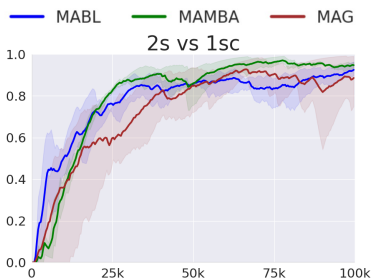

Figure 7: Comparisons against CTCE baselines on 2s vs 1sc

## B.5 Conceptual Comparison: Multi-Agent Latent Variable World Models

| Algorithm | Global information as observations of all agents | Global information other than/in addition to observations of all agents | Decentralized execution | Shared model parameters |
|---|---|---|---|---|
| MABL | ✓ | ✓ | ✓ | ✓ |
| Dreamer-v2 | ✗ | ✗ | ✓ | ✓ |
| MAMBA | ✓ | ✗ | ✗ | ✓ |
| MAG | ✓ | ✗ | ✗ | ✗ |

Table 6: Conceptual comparison of MABL, model-based CTDE and CTCE baselines